\documentclass[letterpaper, 10 pt, conference]{ieeeconf}
\usepackage{graphicx} % Required for the inclusion of images
\usepackage{hyperref}
\usepackage{subcaption}
\usepackage{colonequals}
\usepackage{amsmath} % Required for some math elements
\usepackage{float}
\usepackage{booktabs}
\usepackage{amsfonts}
\usepackage{amssymb}
\usepackage[dvipsnames]{xcolor}
\usepackage{siunitx}
\usepackage{ulem}
\usepackage[font=small,labelfont=bf]{caption}
\DeclareMathOperator*{\argmax}{arg\,max}

\usepackage[ruled]{algorithm2e}
\makeatletter
\newcommand{\algorithmstyle}[1]{\renewcommand{\algocf@style}{#1}}
\newcommand{\removelatexerror}{\let\@latex@error\@gobble}
\newcommand{\norm}[1]{\left\lVert#1\right\rVert}
\makeatother
\usepackage[noadjust]{cite}

\IEEEoverridecommandlockouts                              % This command is only needed if 
\title{\LARGE \bf
	Visual Geometric Skill Inference by Watching Human Demonstration
}

\author{Jun Jin$^{\dagger}$, Laura Petrich$^{\dagger}$, Zichen Zhang$^{\dagger}$, Masood Dehghan$^{\dagger}$ and Martin Jagersand$^{\dagger}$% <-this % stops a space
\thanks{$^{\dagger}$Authors are with the Department of Computing Science,
        University of Alberta, Edmonton AB., Canada, T6G 2E8.
        { 
           \tt\small \{jjin5,laurapetrich,zichen2,masood1,mj7\} @ualberta.ca
        }
        }%
}

\begin{document}
	\maketitle
	\thispagestyle{empty}
	\pagestyle{empty}
	
	%Learning from Demonstration; Visual Learning; Visual Servoing; 
	%222947 179524  1066990 142612
	%%%%%%%%%%%%%%%%%%%%%%%%%%%%%%%%%%%%%%%%%%%%%%%%%%%%%%%%%%%%%%%%%%%%%%%%%%%%%%%%
	\begin{abstract}
		We study the problem of learning manipulation skills from human demonstration video by inferring the association relationships between geometric features. Motivation for this work stems from the observation that humans perform eye-hand coordination tasks by using geometric primitives to define a task while a geometric control error drives the task through execution. We propose a graph based kernel regression method to directly infer the underlying association constraints from human demonstration video using Incremental Maximum Entropy Inverse Reinforcement Learning (InMaxEnt IRL). The learned skill inference provides human readable task definition and outputs control errors that can be directly plugged into traditional controllers. Our method removes the need for tedious feature selection and robust feature trackers required in traditional approaches (e.g. feature-based visual servoing). Experiments show our method infers correct geometric associations even with only one human demonstration video and can generalize well under variance.
	\end{abstract}
	
	%%%%%%%%%%%%%%%%%%%%%%%%%%%%%%%%%%%%%%%%%%%%%%%%%%%%%%%%%%%%%%%%%%%%%%%%%%%%%%%%
	\section{Introduction}
	\label{sec:intro}

	Understanding and applying the mechanism of learning by watching has been researched in robotics for over two decades\footnote{The earliest work can be traced back to Ikeuchi et al.~\cite{ikeuchi1994toward} and Kuniyoshi et al.~\cite{kuniyoshi1994learning} in 1994.}, where the core problem is how to extract high-level reusable symbolic task definitions by observing a human demonstration~\cite{ikeuchi1994toward,kuniyoshi1994learning}. Most of the research focuses on learning task goal configurations rather than task execution~\cite{ahmadzadeh2015learning,dehghan2018online}. This approach reduces the learning complexity and, most importantly, extracts an abstract task representation which allows for generalization. Symbolic task plans can be represented as a tree~\cite{yang2015robot} or graph structure~\cite{shukla2015unified,xiong2016robot} based on the assumption that a task can be decomposed into low-level conditioned elementary skills~\cite{dillmann1995acquisition}, such as, grasping, striking~\cite{kober2010movement}, alignment~\cite{jin2018robot}, and peg-in-hole~\cite{schoettler2019deep}. In order to define the symbols~\cite{ahmadzadeh2015learning}, (action, object, task) recognition techniques and a predefined skill sub-module are hand engineered~\cite{ahmadzadeh2015learning}. These predefined manipulation skills are highly task-dependant and do not generalize well in practice.
	
	\begin{figure}[tbp]
	\setlength{\belowcaptionskip}{-10pt}
		\begin{center}
			\includegraphics[width=0.48\textwidth]{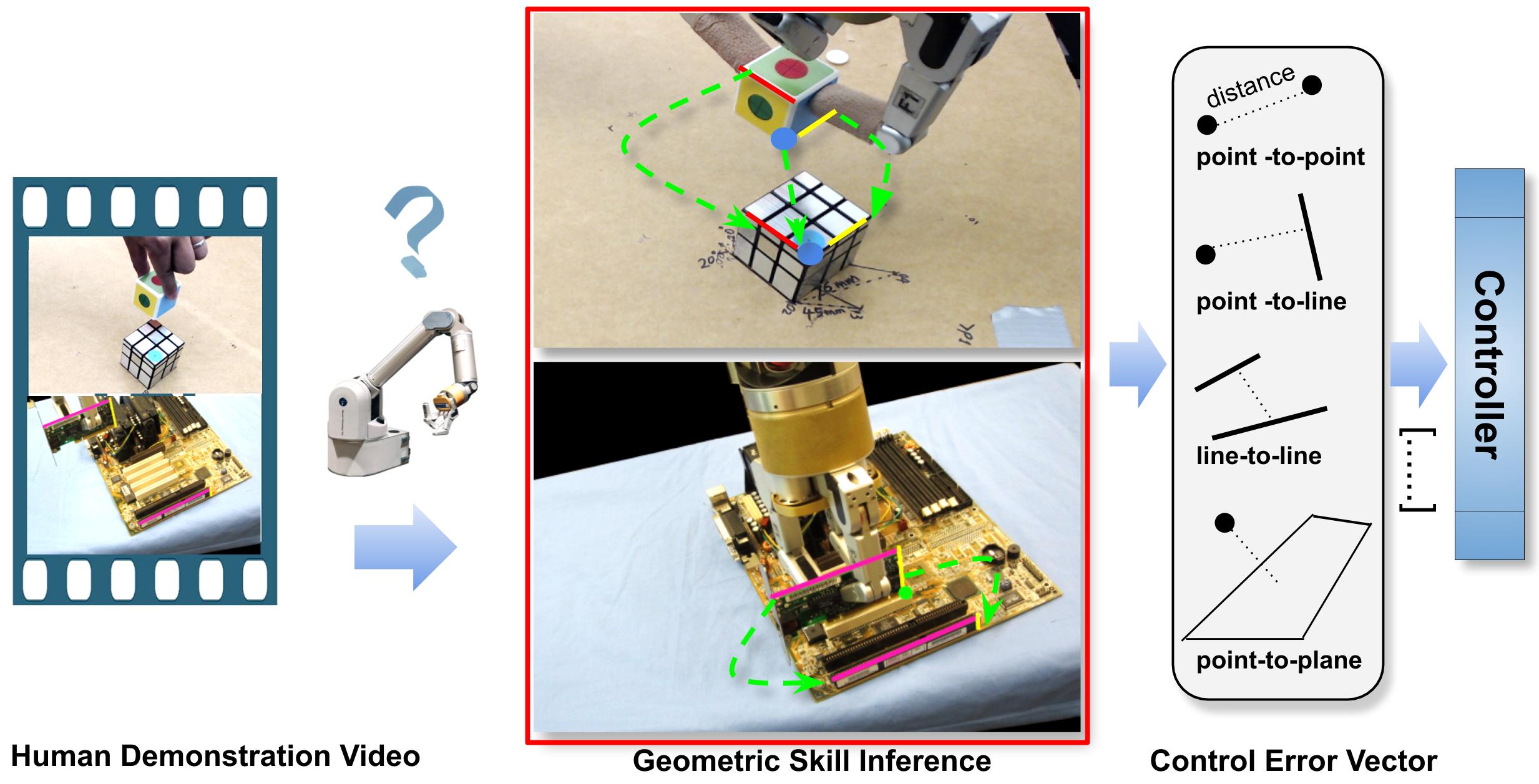} % Include the image placeholder.png
			\caption{Representation of manipulation skills using constraints association between geometric primitives. For example, the alignment skill (or insertion skill) is a combination of several point-to-point or co-linearity constraints; This parameterization partitions the problem into two parts: the geometric task representation and its control error output by computing the Euclidean distance between geometric primitives.
			}
			\label{fig:design_overview}
		\end{center}
	\end{figure}
	
	The main question is whether a general solution exists to parameterize a task. There is no absolute answer, but even if a parameterization exists, it is difficult to find because manipulation tasks are too complex in general. One way of addressing this problem is to use low-level elementary skills as the corner stones of a task; these are potentially easier to learn and generalize. Among the various types of skills, we are interested in those that can be {generally parameterized using geometric association constraints} (Fig. \ref{fig:design_overview}), since a variety of skills can be created from their combinations. We name these \textbf{geometric skills}, which are inherently represented in the image space by geometric primitives (points, lines, conics, planes, etc.) and similar to how human eye-hand coordination works~\cite{hutchinson1996tutorial}. This parameterization method was introduced by Dodds et al. ~\cite{dodds1999task} to solve the box-packing task and then implemented by Gridseth et al. ~\cite{Gridseth2016} on various skills, including, grasping, placing, insertion, and cutting.
	
	This approach, however, has several drawbacks. The task specification is tedious, as it requires manual assignment of associations among geometric features, and is highly dependent on robust feature trackers~\cite{Bateux2017}. This paper aims to address these issues by learning from watching. We propose a method to directly \textit{regress the geometric association constraints} on each frame. 
	
	The main contributions of this paper are:
	\begin{itemize}
		\item Provide an interpretable and invariant robotic task representation using geometric features and their association constraints that are easy to monitor and validate.
		\item Remove the dependency on robust feature trackers in previous methods~\cite{Dodds1999,Gridseth2016} by directly estimating the association constraint between geometric features.
		\item Provide a robust feature association learning method by utilizing the fact that multiple feature associations can define the same task.
	\end{itemize}
	
	To elaborate details of our third contribution, for example, when some features are occluded, other candidates will make up for this and continue to define the task. In contrast, traditional tracking-based methods fix such associations in the initial feature selection stage and may be unable to recover from occlusions. This stiffness on constraints is removed in our maximum entropy-based geometric constraint regression method. 
	
	The remainder of this work will focus on two issues: (1) how to generally encode different types of geometric association constraints and build more complex geometric skills from them; and (2) how to optimize such constraints given one human demonstration video. Experimental results are reported in Sec.~\ref{sec:experiments}.
	
	\begin{figure*}
		\centering
		\includegraphics[width=0.98\textwidth]{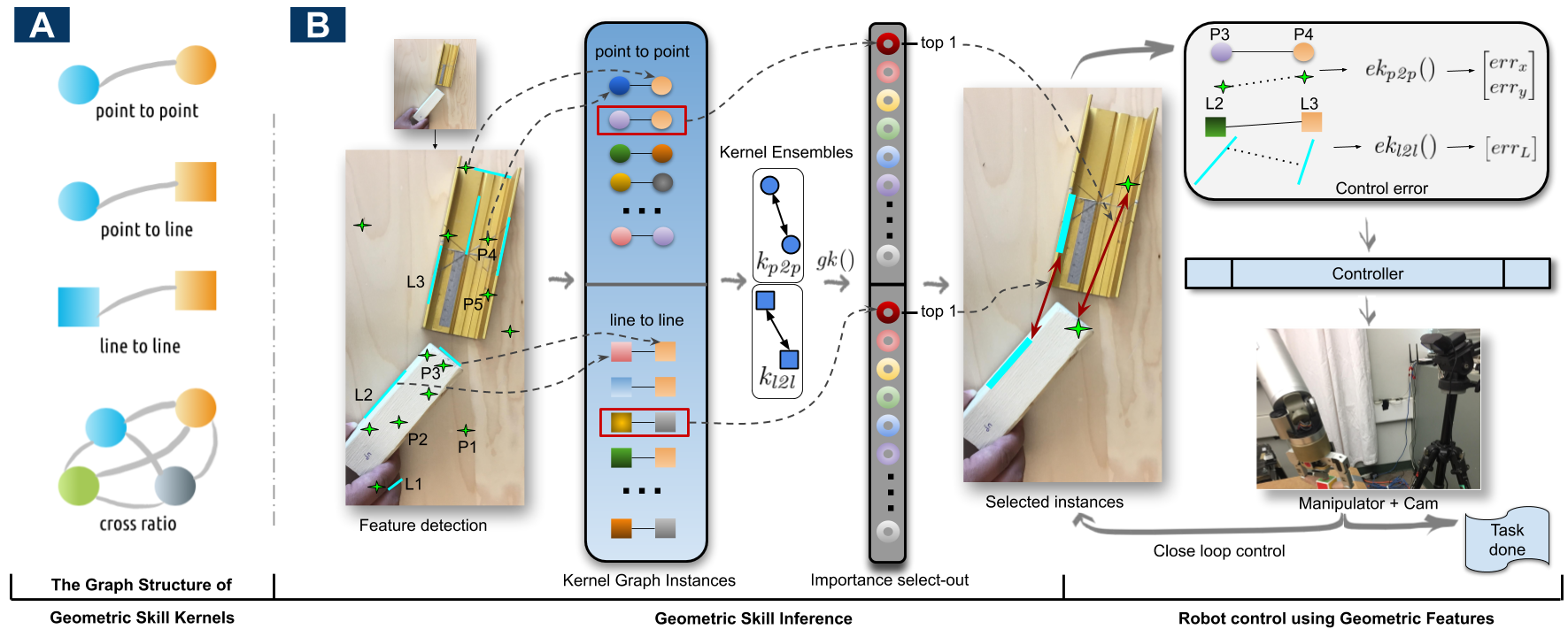}
		\caption{\textbf{A}: Graph structured skill kernels. \textbf{B}: Function of a skill kernel and kernel ensembles. A skill kernel takes input of all geometric feature association instances (kernel graph instances) and output their rank of relevance (select-out) w.r.t. the skill definition (defined by human demonstration video). A skill is a combination (kernel ensembles) of several skill kernels. For example, an `insertion' skill consists of a point-to-point $\mathit{k_{p2p}}$ and a line-to-line $\mathit{k_{l2l}}$ skill kernel. Given one image observation, we can enumerate all possible geometric feature associations. By feeding their corresponding descriptors \{$f_{i}$\}, each association will create one kernel graph instance and each kernel instance will output a select-out to decide which association should be selected. A control error computed from their corresponding \{$y_{i}$\}.}
		\label{fig:diagram}
	\end{figure*}
	
	\section{Related Works}
	\label{sec:review}
	This paper is inspired by research works in robot learning, visual servoing and graph-based relational inference.
	
	\textbf{End to end learning by watching}: This approach, commonly known as imitation learning~\cite{Abbeel2004}, has been recently gaining interest. Sermanet et al. presented TCN~\cite{sermanet2018time} to learn from contrastive positive and negative frame changes along time. Yu et al. proposed a meta-learning based method~\cite{yu2018one} to encode prior knowledge from a few thousand human/robot demonstrations, then learned a new task from one demonstration. End to end learning approaches lack interpretability. Furthermore, to the authors' knowledge, learning by watching only from one human demonstration is still difficult.
	
	\textbf{Learning task plans by watching}: This approach provides the most intuitive motivation and contributes to many of the early works in learning by watching. Such approaches try to generate human readable symbolic representations at a semantic level~\cite{Dillmann2004,yang2015robot,xiong2016robot} to provide high level task planning, which is important for generalizability. Ikeuchi et al. presented a general framework~\cite{ikeuchi1994toward} that relies on object/task/grasp recognition to generate assembly plans from observation. Modern approaches use a grammar parser~\cite{yang2015robot}, causal inference~\cite{xiong2016robot}, and neural task programming~\cite{xu2018neural}. Konidaris et al. proposed constructing skill trees~\cite{konidaris2012robot} at the trajectory level to acquire skills from human demonstration using hierarchical reinforcement learning (RL) with options. This work presents a general framework to learn a tree level structured task. However, such works require hard-coded recognition submodules or lacks generality in various tasks.
	
	\textbf{Learning correspondence relationships by watching}: Learning correspondence relationships to represent a task concept from human demonstration videos provides a generalizable task representation. Current approaches formulate the correspondence relationship through learning at either the \textit{object level}~\cite{florence2018dense} or \textit{key points level}~\cite{manuelli2019kpam, qin2019keto}. Beyond a simple correspondence relationship representation, Sieb et al. propose a graph-structured object relationship inference method~\cite{sieb2019graph} in visual imitation learning. However, apart from learning relationships in the objects or key points level, using a more general framework to construct complex tasks from constraints among fine-grained geometric primitives (points, lines, conics) has rarely been studied.
	
	\textbf{Geometric approaches in skill learning}: Constructing skills using geometric features provides good interpretability. Apart from works mentioned in Sec. \ref{sec:intro}, Ahmadzadeh et al. proposed a system called VSL~\cite{ahmadzadeh2015learning} that is capable of learning skills from one demonstration. VSL first detects objects in an image and represents them using image feature extractors like SIFT. It computes object spatial motion changes via feature matching and then forms a new task goal configuration used to generate motion primitives by a trajectory-based learning from demonstration (LfD) method~\cite{ijspeert2013dynamical}. Landmark-based pre/post action condition detection is also used to construct a task plan. Triantafyllou et al. proposed a geometric approach to solve the garment unfolding task~\cite{triantafyllou2016geometric}. Tremblay et al. proposed a human-readable plan generation method~\cite{Tremblay2018} which provides interpretability by modeling the 3D wire frame of blocks, however, it requires simulator training for prior 3D modelling.
	
	\section{Method}
	\label{sec:method}
	\subsection{Geometric Skill Kernels}
	Let $\mathcal{O}$ denote the observation space and $\mathcal{F}$ denote the observed geometric features\footnote{points, lines, conics, planes, spheres etc. from an image or point cloud.}. Each feature has two parts: a descriptor $f_{i}$ that encodes locally invariant properties and a coordinate parameter set $y_{i}$ that encodes globally geometric properties\footnote{More details on the parameterization of geometric primitives in~\cite{hartley2003multiple}.}. 
	A geometric skill kernel $\mathit{k}$ is a composite functional structure that describes association constraints between geometric features $(f_{i}, y_{i}) \in \mathcal{F}$. To ground our formalism, we describe some basic examples:
	\begin{itemize}
		\item \textit{point-to-point} $\mathit{k_{p2p}}$: the coincidence of two points.
		\item \textit{point-to-line} $\mathit{k_{p2l}}$: a point is on a line.
		\item \textit{line-to-line} $\mathit{k_{l2l}}$: a line is collinear with another line.
		\item \textit{coplanarity} $\mathit{k_{copl}}$: coplanar four points or two lines.
	\end{itemize}
	
	Each kernel has two parts: a geometric association constraint representation part and a control error generation part used to guide robot actions.
	
	\subsubsection{{Geometric association constraint representation}}\label{ambiguity}
	Inspired by graph motifs~\cite{zellers2018neural}, each skill kernel is a unit graph with different structures. An undirected graph $\mathcal{G}=\{V, E\}$ is used to represent the association constraint, where nodes $V$ are variables that take input of feature descriptors $\{f_{1},...,f_{n}\}$, and edges $E$ define a fixed graph structure (as shown in Fig. 2A). For example, the graph for $\mathit{k_{p2p}}$ has two connected nodes, and each node $v_{i}$ corresponds to $f_{i}$. By feeding in two points, we get a graph instance $\mathcal{G}$ and use a select-out function $\mathit{gk}$ to measure how relevant it is to define the skill. Then, we have:
	\begin{equation}
	\label{eq:kernel}
	\mathit{gk}: \mathcal{G} \xrightarrow{} [0,1]\in \mathbb{R}, \;\; \mathit{gk}(\mathcal{G}(\{f_{1},...,f_{n}\}))
	\end{equation}
	For example, in the `insertion' skill (Fig. 2B), the graph instance of P3 and P4 has higher $\mathit{gk}$ output than that of P1 and P2, and will be selected out. 
	
	It is worth noting the \textit{ambiguity property} of skill kernels, where several association instances may define the same skill. Because of this property, it is crucial to learn a robust feature association selection model in the long-run since when the current association is not available (occluded or outside of the field of view (FOV), a candidate association will be selected. For example, in Fig. 2B, both the association of P3 to P4 and P2 to P5 can partially define the skill. In successive steps, P5 will be occluded so its instance won't be observable, however, P3 to P4 can make up the role. Another issue is task \textit{decidability}, which determines which 2D image-coordinate constraints are needed to guarantee a particular 3D configuration. We do not cover decidability here, but refer to ~\cite{dodds1999task}.
	
	\subsubsection{{Geometric control error generation}}
	Let $\mathit{E_k}: \{y_{1},...,y_{n}\} \xrightarrow{} \mathbb{R}^{d}$ denote the mapping of all nodes geometric parameters to a control error vector where $d$ is the degree of freedom that this constraint contributes. For example, given a \textit{point-to-point} skill, $d=2$ for image points. The control error of a point-to-point kernel is the point distance, while of a point-to-line kernel is the dot product of their homogeneous coordinates. More examples can be found in~\cite{Dodds1999,Gridseth2016}. $\mathit{E_k}$ will be used in the following optimization using human demonstrations and in generating control signals used to guide robot action.
	
	\subsection{Parameterization}
	\subsubsection{Parameterization of $\mathcal{G}$}
	$\mathcal{G}$ is parameterized by a $T$-layer message passing graph neural network~\cite{gilmer2017neural}. Each node $v_{i} \in \mathcal{G}$ relates a $h$-dimensional hidden state $h^{i}_{t}$. At layer $t$ (or time step $t$), each nodes hidden state $h^{i}_{t}$ is updated via three steps. (I) Pair-wise message generation $\mathcal{M}$:
	\begin{equation}
	\begin{aligned}
	m^{t+1}_{i \rightarrow j}=\mathcal{M}(h^{t}_{i}, h^{t}_{j})
	\end{aligned}
	\end{equation} 
	where $h^{t}_{j}$ relates to any node $v_{j}$ connected to $v_{i}$. (II) Message aggregation $\mathcal{A}$ which collects all incoming messages:
	\begin{equation}
	\begin{aligned}
	m^{t+1}_{i}=\mathcal{A}(m^{t+1}_{j \rightarrow i})
	\end{aligned}
	\end{equation} 
	We simply use summation as $\mathcal{A}$ in our implementation. Lastly, (III) message update $\mathcal{U}$:
	 \begin{equation}
	 \begin{aligned}
	 h^{t+1}_{i}=\mathcal{U}(h^{t}_{i},m^{t+1}_{i})
	 \end{aligned}
	 \end{equation}
	where a gated recurrent unit (GRU) is used. After $T$ layer updates, all of the nodes final states are fed into a \textit{MLP} layer with an activation function that outputs a scalar value $b=\sigma (\text{MLP}(h^{T}_{1},...,h^{T}_{n}))$.
	
	%MJ why is this called "select-out" ionstead of just "select"?
	\subsubsection{The select-out function $\mathit{gk}$}
	Given one image, we construct $m$ graph instances by enumerating all possible geometric primitive combinations (e.g., point-to-point by listing association between any two points). Each instance \{$\mathcal{G}_{i}$\} represents one association and will output its relevance $b_{i}$. A select-out function $\mathit{gk}$ outputs a relevance factor $g_{i}$:
	\begin{equation}
	\begin{aligned}
	g_{i}=\mathit{gk}(\mathcal{G}_{i})=\text{softmax}(b_{i},\{b_{1},...,b_{m}\})
	\end{aligned}
	\end{equation}
	
	Let $\mathit{E_{k}}^{i}$ denotes the control error for each graph instance, we now define the overall control error $\mathbf{Ec}$ for a whole image as:
	\begin{equation}
	\begin{aligned}
	\mathbf{Ec}=\sum_{i=1}^{m} g_{i}\mathit{E_{k}}^{i}
	\end{aligned}
	\end{equation}
	
	\subsection{InMaxEnt IRL for optimization}
	Given human demonstration video frames, we apply \textit{InMaxEnt IRL}~\cite{jin2018robot} for optimization. To this end, we define the reward function, which connects skill kernels to entropy models. Through the optimization of this reward function, the skill kernel is also optimized. In practice, each skill kernel is optimized individually. We use $\mathit{k_{p2p}}$ as an example in the following discussion. 
	
	\begin{figure}[tbp]
	\setlength{\belowcaptionskip}{-10pt}
		\begin{center}
			\includegraphics[width=0.45\textwidth]{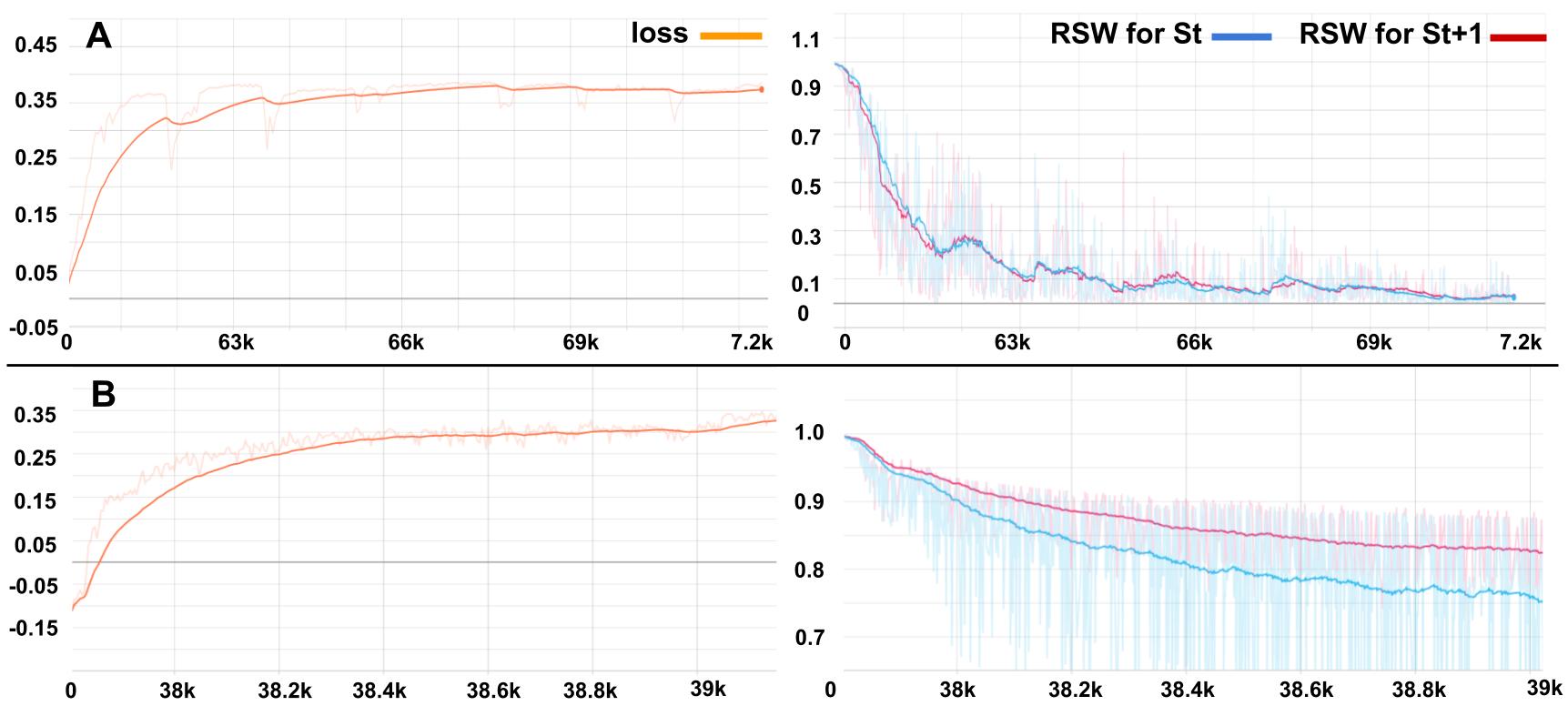} % Include the image placeholder.png
			\caption{RSW measures how much relevance contributed from the non-selected association instances. The lower RSW, the more deterministic in select-out. \textbf{A} shows the training curve with RSW regularizer. The relevance from remaining instances stays below 0.1\%. \textbf{B} is without RSW regularizer. Although the cost function is optimized, the non-selected ones still occupy 75\% of relevance. Note that we are maximizing the loss.}
			\label{fig:rsw}
		\end{center}
	\end{figure}
	
	\subsubsection{Reward function}\label{reward}
	Each state $s_{t}$ is an image related to a control error $\mathbf{Ec}_{t}$, where  the subscript  $t$  denotes the time step in RL. An optimized $\mathit{k_{p2p}}$ should \textit{consistently} select \textit{`correct instances'} among all states. During human demonstration we should expect $\mathbf{Ec}_{t}$ to decrease globally (but not necessarily in each step). Intuitively, we should get a positive reward if we observe a decrease in $\mathbf{Ec}_{t}$, otherwise the reward should be negative. Let $\Delta \mathbf{Ec}_{t}=\norm{\mathbf{Ec}_{t+1}}-\norm{\mathbf{Ec}_{t}}$, we define:
	
	\begin{equation}
	\begin{aligned}
	r_{t}=\frac{2}{{1+exp(\beta \Delta \mathbf{Ec}_{t})}} - 1
	\end{aligned}
	\end{equation}
	$r_t\in (-1,1)$, where $\beta$ normalizes the scale of different skill kernels' output $E_k$.
	
	\subsubsection{The variational expert assumption}
	\textit{InMaxEnt IRL} considers imperfect expert demonstrations with a confidence level $\alpha$. A higher confidence level results in smaller variance $\sigma_{0}$ in demonstration. We assume that in a human demonstration, at state $s_{t}$ the probability of selecting an action that transitions to the observed state $s_{t+1}$ follows a Boltzmann distribution with conditions:
	\begin{equation}
	\begin{aligned}
	p(s_{t+1}|s_{t})=\frac{1}{\mathcal{Z}_{t}}exp(r^{*}_{t})p(r^{*}_{t}),
	\end{aligned}
	\end{equation}
	where $r^{*}_{t}$ is the reward of this observed state change, and 
	\begin{equation}
	\begin{aligned}
	\mathcal{Z}_{t}= \mathbb{E}_{p(r_{tj};r^{*}_{t})}[exp(r_{tj})]
	\end{aligned}
	\end{equation}
	is the partition function, $r_{tj} \sim \mathcal{N}(r_{t}^{*}, \sigma_{0} ^{2})$ is a truncated normal distribution with domain in [-1,1]. This means that the expert prefers the action with the highest reward among all possible actions $\mathcal{A}_{t}=$\{$a_{tj}$\}. To emphasize high impact actions in $\mathcal{A}_{t}$, suppose $a_{tj}$ gets a reward $r_{tj}$, the chance of $a_{tj}$ included in the pool is: $p(r_{tj}) = \mathcal{N}(r_{t}^{*}, \sigma_{0})$, this is called a human factor~\cite{jin2018robot} since it varies with the human demonstrator's confidence. 
	
	\subsubsection{Loss function}
	To maximize the probability of observed human demonstration video sequence $p(\{s_{t}\})$ by applying MDP property, we have:
	\begin{equation}
	\label{eq:cost}
	\mathcal{L}  = \argmax_{\theta} \sum {log[p(s_{t+1}|s_{t})]}
	\end{equation}
	With equation (8) and removing the last constant, the cost function can be further written as:
	\begin{equation}
	\label{eq:loss}
	\mathcal{L}  = \argmax_{\theta} \sum r^{*}_{t} - log \mathcal{Z}_{t}
	\end{equation}
	Note that if $p(r_{tj})$ has domain $(-\infty ,\infty)$, the loss function is a constant. Proofs can be found on our \href{http://webdocs.cs.ualberta.ca/~vis/Jun/InMaxEntIRL/supplymentary-material.pdf}{website}~\cite{proofs}.
	
	 To force $\mathit{gk}$ into making selections more deterministic meanwhile considering the \textbf{ambiguity property}, a penalty regularizer $-\lambda \, RSW$ is added to the reward where $\lambda$ is a hyperparameter and \textit{RSW} is the residual sum of weights (\textbf{RSW}). This makes $gk$ output major weights on selected $p$ alternatives while minimizing the residual sum of weights. Fig. 3 shows a comparison of training with and without RSW penalty in the \textit{Sorting} task.
		%% finally the full algorithm
	\begingroup
	\removelatexerror% Nullify \@latex@error
	{\small
	\begin{algorithm}[tbp]
		\SetAlgoLined
		\KwIn{Expert demonstration video frames \{$s_{1},...,s_{n}$\}, confidence level $\alpha$}
		\KwResult{Optimal weights $\boldsymbol{\theta} ^{*}$ of $\mathit{k_{p2p}}$}
		\textcolor[rgb]{0.14,0.36,0.73}{\textbf{Construct kernel graph instances on each frame}}\\
		\For{t = 1:n}{
			Feature point extraction on $s_{t}$ to get \{$(f_{i}, y_{i})$\}\\
			Enumerate all $\mathit{k_{p2p}}$ instances by association\\
			Feed all instances to $\mathit{gk}$ to get $\mathbf{Ec}_{t}$\\
		}
		\textcolor[rgb]{0.14,0.36,0.73}{\textbf{Prepare State Change Samples $\mathcal{D}{s}=s_{t}\rightarrow s_{t+1}$}}\\
		Compute $\sigma_{0}$ using $\alpha$; \textcolor[rgb]{0.14,0.36,0.73}{\textbf{Shuffle $\mathcal{D}{s}$; Initialize $\boldsymbol{\theta} ^{0}$}}\\
		\For{each iteration}
		{
			\For{each observed sample change in $\mathcal{D}{s}$}{
				\textcolor[rgb]{0.14,0.36,0.73}{\textbf{Forward pass}}\\
				Compute $r_{t}^{*}$\\
				Compute $\nabla _{r_{t}^{*}}\mathcal{L}=\sum 1 - \frac{1}{\mathcal{Z}_{t}} \nabla _{r^{*}_{t}} \mathcal{Z}_{t}$\\
				$grad = \mathit{k_{p2p}}.backProp(\nabla _{r_{t}^{*}}\mathcal{L})$\\
				\textcolor[rgb]{0.14,0.36,0.73}{\textbf{Gradient ascent update}}\\
				$\boldsymbol{\theta} ^{n+1}=updateWeights(\boldsymbol{\theta} ^{n}, grad)$
			}
		}
		
		\caption{Optimizing $\mathit{k_{p2p}}$}
	\end{algorithm}
	}
	\endgroup
	
	\subsubsection{Optimization}
	The last item in eq. (11) is a constant and $\mathcal{Z}_{t}$ is a function of $r^{*}_{t}$, which is further represented using skill kernels with parameters $\theta$. Then, we have:
	\begin{equation}
	\label{eg:opt}
	\nabla _{\theta}\mathcal{L} = \sum \nabla _{\theta} r^{*}_{t} - \frac{1}{\mathcal{Z}_{t}} \nabla _{r^{*}_{t}} \mathcal{Z}_{t} \nabla _{\theta} r^{*}_{t}
	\end{equation}
	
	$\nabla _{\theta} r^{*}_{t}$ can be solved by back propagation from eq. (7) to the graph neural network in the skill kernel. $\mathcal{Z}_{t}$ can be estimated by a Monte Carlo estimator sampling $s_1$ samples from the truncated normal distribution $p(r_{tj})$:
	
	\begin{equation}
	\label{eq:Z}
	\mathcal{Z}_{t} \approx \frac{1}{s_1}\sum^{s_1} exp(r_{tj}),\:\: r_{tj} \sim p(r_{tj})
	\end{equation}
	$\nabla _{\theta} \mathcal{Z}_{t}$ is the derivative of an expectation. By applying the log derivative trick, we have:
	\begin{equation}
	\begin{aligned}
	\nabla _{r^{*}_{t}} \mathcal{Z}_{t}&=\nabla_{r^{*}_{t}}\mathbb{E}_{p(r_{tj})}[exp(r_{tj})]\\
	&=\mathbb{E} _{p(r_{tj})}[exp(r_{tj})\nabla_{r^{*}_{t}} log \,p(r_{tj})]\\
	&\approx\frac{1}{s_2}\sum^{s_2} exp(r_{tj})\nabla_{r^{*}_{t}} log \,p(r_{tj}),\:\:\ r_{tj} \sim p(r_{tj})
	\end{aligned}
	\end{equation}
	Since $p(r_{tj})$ is tractable, it's trivial to get:
	\begin{equation}
	\begin{aligned}
	\nabla_{r^{*}_{t}} log \,p(r_{tj})=\frac{x_{\mu}}{\sigma_{0}} + \frac{exp(-b_{\mu}^{2}/2)-exp(-a_{\mu}^{2}/2)}{\sqrt{2\pi}\sigma_{0}[\phi(b_{\mu})-\phi(a_{\mu})]}
	\end{aligned}
	\end{equation}
	$x_{\mu}=(r_{tj}-r_{t}^{*})/\sigma_{0}$, $a_{\mu}=(-1-r_{t}^{*})/\sigma_{0}$, $b_{\mu}=(1-r_{t}^{*})/\sigma_{0}$.
	where $\phi$ is defined in~\cite{truncnorm}. By combining the above equations, $\nabla _{\theta}\mathcal{L}$ is solved. 
	
	The optimization on $\mathit{k_{p2p}}$ is summarized in Algorithm 1. 

	\subsection{From Skill Kernel to Skills}
	In this paper, we consider that a skill is simply the combination of several skill kernels, namely \textit{kernel ensembles}. There should be more advanced ways to construct a skill from different kernels, although this is not discussed here. \footnote{For example, for a `peg-in-hole' skill, the point-to-point kernel should be used to first coarsely move to the target, while line-to-line kernel best fits in the final alignment actions. Their relationship is not a simple combination.}
	
	\subsection{From Skill to Control}
	Given an image, each skill kernel will select out several alternative association instances and generate control error vectors. For example, the point-to-point $\mathit{k_{p2p}}$ will output vectors with structure [$err_{x}, err_{y}$], which can be plugged in controllers like feature-based visual servoing or uncalibrated visual servoing~\cite{jagersand1995visual}. More examples using geometric features (lines, conics) and based on which, the constructed skill kernels in UVS control are included in~\cite{Gridseth2016}.

\begin{figure}[h]
	\setlength{\belowcaptionskip}{-10pt}
		\begin{center}
			\includegraphics[width=0.3\textwidth]{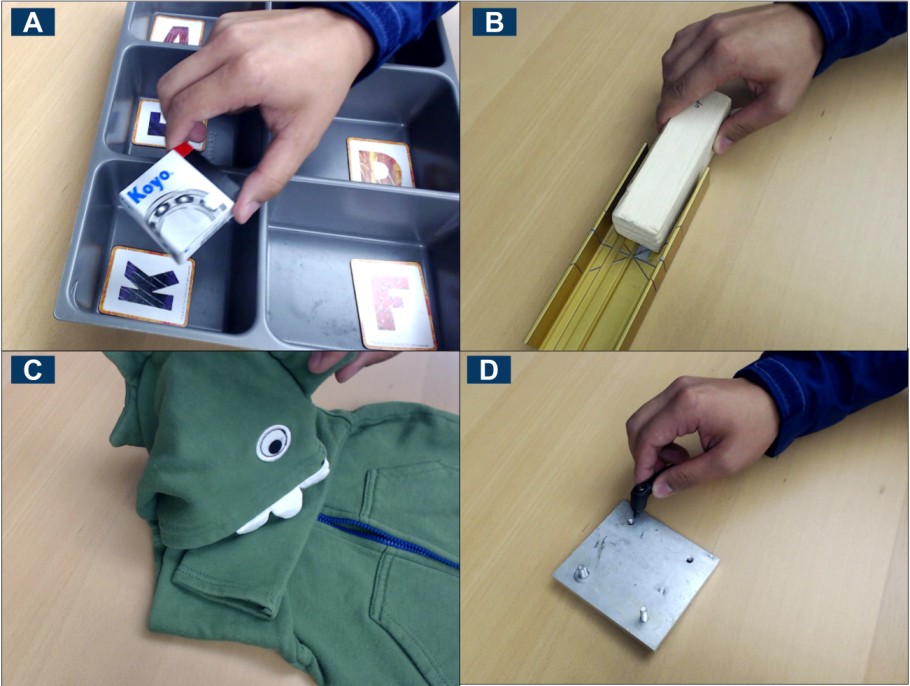} % Include the image placeholder.png
			\caption{Four types of skills with human demonstration. A: \textit{Sorting} skill. B:\textit{Insertion} skill. C:\textit{Folding} cloth skill. D: Driving a \textit{Screw} to the hole skill.}
			\label{fig:skills}
		\end{center}
	\end{figure}

    \section{Experiments}
	\label{sec:experiments}
	\subsection{Quantitative Evaluation}
	We first evaluate what types of skills the learned inference behavior is capable of. Four types are tested (Fig.~\ref{fig:skills}): \textit{Sorting} skill represents a regular setting; \textit{Insertion} is for skills that need line-to-line constraint; \textit{Folding} is for manipulation with deformable objects; and \textit{Screw} skill represents types that have low image textures. Each skill is evaluated on videos that show a human performing the same task but with random behaviors. The objective is to infer the correct geometric feature associations that can be used to define the demonstrated skill.

	We next {test if the learned behavior from human demonstration video directly generalizes to a robot hand.} For our tests, the background table is also changed and the target pose is randomly arranged (Fig.~\ref{fig:setup}A). 
	
	Lastly, we test on the robot with four other scenarios (Fig.\ref{fig:exp1}, B-E): moving camera; occlusion; object running out of camera's field of view; and illumination changes.
	
	\begin{figure}[h]
	\setlength{\belowcaptionskip}{-10pt}
		\begin{center}
			\includegraphics[width=0.35\textwidth]{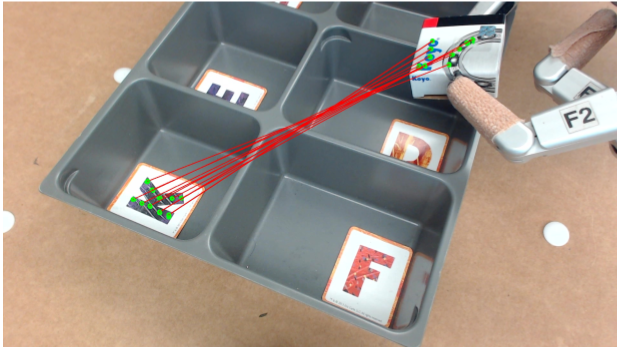} % Include the image placeholder.png
			\caption{The hand designed baseline requires human to specifically select 10 pairs of feature points to define the demonstrated skill.}
			\label{fig:baseline}
		\end{center}
	\end{figure}
	
\textbf{Baseline:} To our best knowledge, there are no existing methods that learn geometric feature associations by watching human demonstration. However, for comparison, we hand designed a {baseline} on the \textit{Sorting} skill. The baseline requires a human to manually select 10 pairs of feature points and initialize 20 trackers. Each pair has one point on the object and another on the target. All of the 10 pairs simultaneously define the same skill (Fig.\ref{fig:baseline}), resulting in a robust baseline. In evaluation, as long as one pair still defines the skill, the baseline is marked as a successful trial.

	\begin{figure}[h]
	\setlength{\belowcaptionskip}{-10pt}
		\begin{center}
			\includegraphics[width=0.48\textwidth]{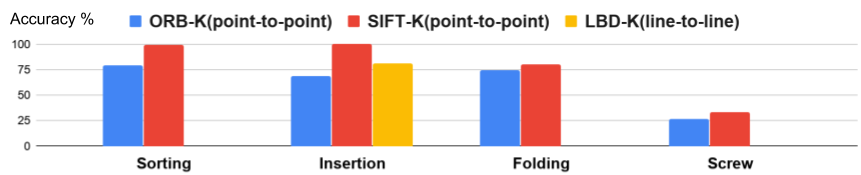} % Include the image placeholder.png
			\caption{Results of the four skills.}
			\label{fig:sift}
		\end{center}
	\end{figure}
\begin{figure}[h]

\setlength{\belowcaptionskip}{-10pt}
		\begin{center}
			\includegraphics[width=0.3\textwidth]{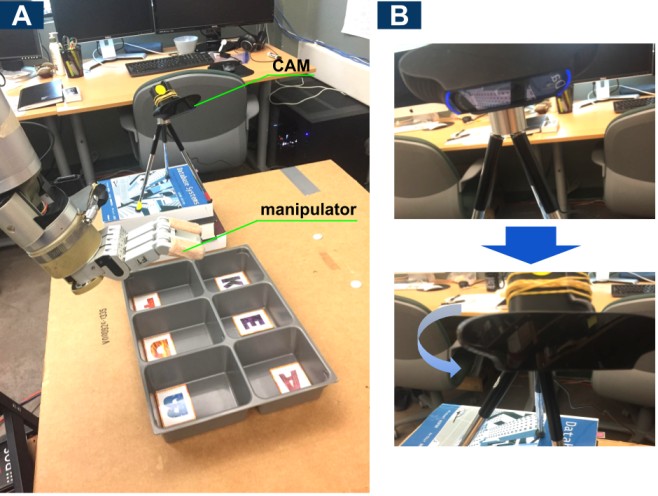} % Include the image placeholder.png
			\caption{\textbf{A}: Experimental setup on the manipulator. \textbf{B}: We change the camera pose in evaluation by rotation and a random displacement.}
			\label{fig:setup}
		\end{center}
	\end{figure}
	
\begin{figure*}[h]
		\centering
		\includegraphics[width=0.9\textwidth]{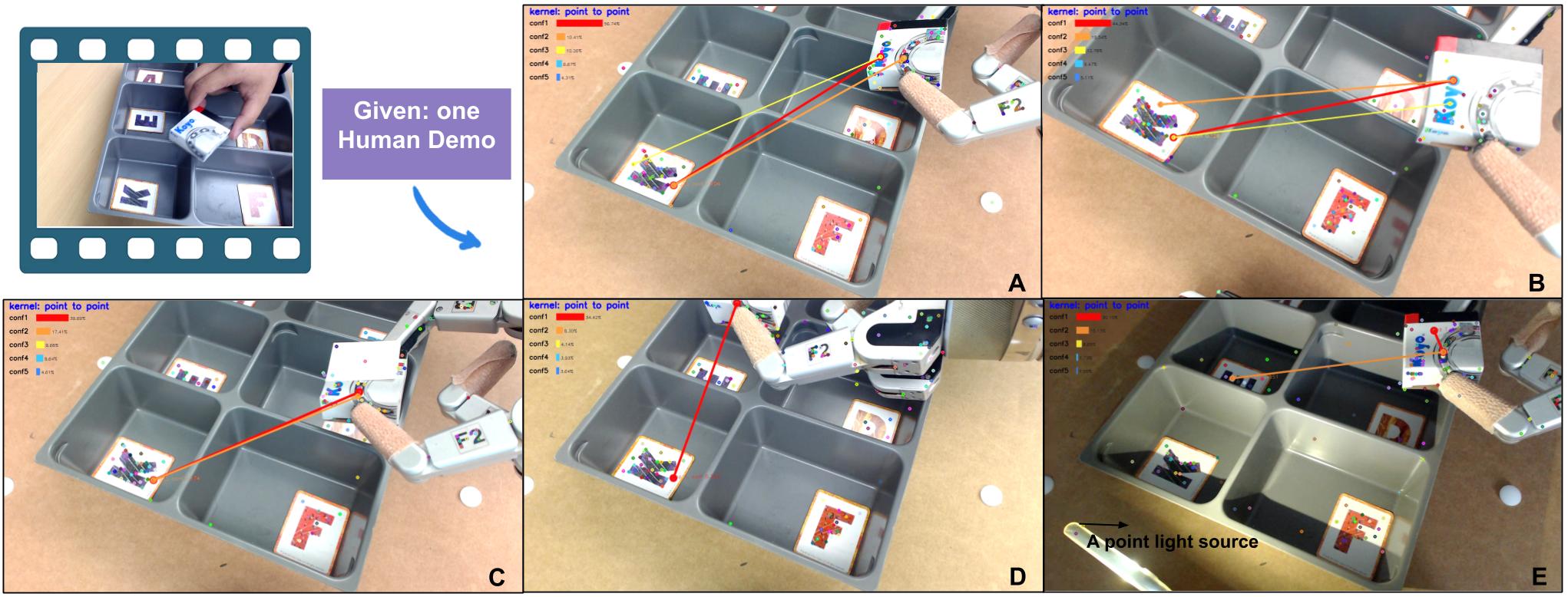}
		\caption{Given one human demonstration video, we evaluate the learned behavior on 5 scenarios. \textbf{A}: using robot hand with a different background and random target pose; \textbf{B}: projective variance due to camera pose change; \textbf{C}: occlusion; \textbf{D}: object out of camera's view-field; and \textbf{E}: illumination change. For each scenario, we detect all feature points and use a colored line to mark the select-out associations. The top one is marked red and the bar next to it indicates the estimation confidence. Only the association with confidence greater than 10\% is displayed. Results are reported in Table~\ref{tab:expB}.}
		\label{fig:exp1}
	\end{figure*} 
	
% 	\begin{figure}[h]
% 	\setlength{\belowcaptionskip}{-10pt}
% 		\centering
% 		\includegraphics[width=0.45\textwidth]{fig5_new.jpg}
% 		\caption{Training curve and reward output from selected associations in the \textit{sorting} skill. \textbf{A}: using SIFT descriptors. \textbf{B}: using ORB descriptors. Results show that using SIFT descriptors exhibits smoother learning. The association instance selections of both are stabilized when the loss converges and output control error stabilizes at the same level.}
% 		\label{fig:sortingskill}
% 	\end{figure}

\textbf{Metric:} We evaluate on each video frame and calculate the accuracy of inferences. For the baseline, when it fails on one frame, it can't be resumed unless a human hand select the features again, therefore we report only success or failure on the final result. For our method, failures can be automatically corrected in successive frames. While our method can output $p$ inferred associations, we pick the top one for evaluation.
	
\subsubsection{Training}
For each skill, we evaluate the point-to-point kernel using SIFT and ORB features respectively. For the \textit{Insertion} skill, we add the line-to-line kernel using LBD~\cite{zhang2013efficient} line descriptor. All kernels have the same graph layer size=5 with hidden state dimension=512 and p=10 alternatives (\ref{reward}). In training, we set the regularizer coefficient $\lambda=0.1$, and human factor $\sigma_{0}=0.55$. Each kernel with different descriptors are trained individually. 
% The training curve of the \textit{Sorting} skill is shown in Fig.\ref{fig:sortingskill} as an example.
\begin{table}[h]
\scriptsize
\renewcommand{\arraystretch}{.7}
\begin{center}
\begin{tabular}{cccccc}
% \hline
% Settings & \textit{Random Target} & \textit{Move Camera} & \textit{Object Occlusion} & \textit{Object Outside FOV} & \textit{Change Illumination} \\ 
 & \textit{Random} & \textit{Move} & \textit{Object} & \textit{Outside} & \textit{Change} \\
 & \textit{Target} & \textit{Camera} & \textit{Occlusion} & \textit{FOV} & \textit{Illumination} \\ 
\hline
Baseline & 100.0\%                & 0.0\%                & 0.0\%                     & 0.0\%                       & 0.0\%                        \\
Ours     & 99.1\%                 & 96.7\%               & 92.7\%                    & 81.2\%                      & 0.0\%                        \\ \hline
\end{tabular}
\caption{Evaluations results of running robot under various environmental settings as shown in Fig.~\ref{fig:exp1}. For each variation setting, we count correct geometric association inferences on each frame and calculate the percentage of successful inferences among all the frames during the execution of \textit{Sorting} task.}
\label{tab:expB}
\end{center}

\end{table}
\subsubsection{Results}
\paragraph{Different skills}
Results (Fig.\ref{fig:sift}) on the 4 skills show our method is capable of the \textit{Sorting} and \textit{Insertion} skill and performs moderately in \textit{Folding} and \textit{Screw} skills. In experiments, we observed that when both object and target have rich textures, results improve. This may be from the use of SIFT or ORB that are local descriptors dependent on textures. We can expect further improvement by using other local feature descriptors~\cite{winder2007learning}~\cite{kumar2016learning}. We also find the more features that can be fed into the skill kernel, the better accuracy it performs. Due to our hardware GPU limitation, we can only test using a small number (60 in average) of features. 

\paragraph{Varying environment}
Fig.~\ref{fig:exp1} lists results on various environmental conditions. In general, i) our method is robust to occlusion. When some feature associations are occluded, the selection of others will make it up; and ii) our method exhibits robust behavior so that failure in some frames doesn't affect successive frames since it directly selects the feature associations on each frame. In contrast, the baseline method depends on the initialization of video trackers and continuous tracking. We observe that the learned inference behavior tends to select fixed association instances while showing the flexibility of selecting alternatives when fixed ones are not observable. We also observe that the accuracy is highly related to the capability of SIFT descriptor. It reaches high accuracy under projective variance (B), however, fails under illumination changes (E).

Although results on the robot manipulator show our method can output the correct selections of geometric feature associations which can be directly used in controllers (e.g. uncalibrated visual servoing~\cite{Gridseth2016}), due to resource limitations we did not test with a plug-in controller. We leave this to our future work.

	\section{Conclusion}
	\label{sec:conclusion}
% 	LP: should the issues here be in list form? Might sound better just as a regular paragraph.
% MJ Paragraph format is fine. It can contain a list of conclusions/results in text (just like the 1), 2) for future work.
	We propose a graph based kernel regression method to infer the association relationship between geometric features by watching human demonstrations. The learned skill inference provides human readable task definition and outputs control errors that fit in traditional controllers. Our method removes the dependency on robust feature trackers and tediously hand selection process in traditional robotic task specification. The learned selection model provides a robust feature association behavior under various environmental settings.
	
	Although results are promising, there are issues that need to be further investigated. 1) Consistent control error output: while the result shows that our method tends to select a fixed set of associations, it can't guarantee the selection consistency. One possible solution is to add constraints between frames. 2) Other local feature descriptors~\cite{winder2007learning}~\cite{kumar2016learning} are worth trying for better generalization. 3) The generalization to point cloud geometric primitive needs to be further studied.
	
	\addtolength{\textheight}{-0 cm}   % This command serves to balance the column lengths
	
	%%%%%%%%%%%%%%%%%%%%%%%%%%%%%%%%%%%%%%%%%%%%%%%%%%%%%%%%%%%%%%%%%%%%%%%%%%%%%%%%

	%%%%%%%%%%%%%%%%%%%%%%%%%%%%%%%%%%%%%%%%%%%%%%%%%%%%%%%%%%%%%%%%%%%%%%%%%%%%%%%%

	%%%%%%%%%%%%%%%%%%%%%%%%%%%%%%%%%%%%%%%%%%%%%%%%%%%%%%%%%%%%%%%%%%%%%%%%%%%%%%%%

	%%%%%%%%%%%%%%%%%%%%%%%%%%%%%%%%%%%%%%%%%%%%%%%%%%%%%%%%%%%%%%%%%%%%%%%%%%%%%%%%
	
% 	\clearpage
	
	\bibliographystyle{IEEEtran}
	\bibliography{IEEEabrv,IEEEexample}
	
\end{document}